\pdfoutput=1

\documentclass[11pt]{article}

\usepackage{EMNLP2023}

\usepackage{tipa}

\usepackage{times}
\usepackage{latexsym}
\usepackage{graphicx}
\usepackage{multirow}

\usepackage[T1]{fontenc}

\usepackage[utf8]{inputenc}

\usepackage{microtype}

\usepackage{inconsolata}

\title{Representing and Computing Uncertainty in Phonological Reconstruction}

\author{%
Johann-Mattis List \\ MCL Chair / DLCE \\ University of Passau / MPI-EVA \\ Passau / Leipzig,
Germany
         \And 
Nathan W. Hill \\ Trinity Centre for Asian Studies \\ University of Dublin \\ Dublin, Ireland
\AND
Robert Forkel \\ DLCE \\ MPI-EVA \\ Leipzig, Germany
\And
Frederic Blum \\ DLCE \\ MPI-EVA \\ Leipzig}
\begin{document}
\maketitle
\begin{abstract}
Despite the inherently fuzzy nature of reconstructions in historical linguistics, most scholars do not represent their uncertainty when proposing proto-forms. With the increasing success of recently proposed approaches to automating certain aspects of the traditional comparative method, the formal representation of proto-forms has also improved. This formalization makes it possible to address both the representation and the computation of uncertainty. Building on recent advances in supervised phonological reconstruction, during which an algorithm learns how to reconstruct words in a given proto-language relying on previously annotated data, and inspired by improved methods for automated word prediction from cognate sets, we present a new framework that allows for the representation of uncertainty in linguistic reconstruction and also includes a workflow for the  computation of fuzzy reconstructions from linguistic data.
\end{abstract}

\section{Introduction}
Phonological reconstruction refers to the techniques that linguists use to reconstruct the phonological and phonetic shape of word forms or morphemes in unattested ancestral languages. 
Although the results are inherently provisional (as witnessed by the changes in the fable by \citealt{Schleicher1868} over the last decades, cf. \citealt{Luehr2008}), linguists typically present their results in the form of discrete phonological units, giving the impression of exactitude and rigor. Thus, although phonological reconstructions change with time as our knowledge or our assumptions about a language family change, we typically provide the results as if they were final. 
By focusing on the uncertainty of phonological reconstructions, we aim to provide a new framework by which uncertainty in phonological reconstruction can be 
a) represented (in etymological databases or etymological dictionaries), and
b) computed (from etymological datasets).
Representation and computation have several benefits. On the one hand, improved representations allow for a more refined reconstruction practice that more directly and consistently indicates the weak points in a reconstruction. On the other hand, computing the uncertainty of a given reconstruction system allows scholars to refine their reconstructions by helping them to identify weak points and potential errors in their cognate judgments or correspondence patterns.

The traditional techniques for phonological reconstruction, by which ancestral word forms are reconstructed from observed words with the help of the comparative method, are of crucial importance for historical language comparison. Despite the inherently fuzzy nature of reconstructions, most scholars have so far hesitated to systematically represent their uncertainty when proposing proto-forms (for an exception see \citealt{Baxter2014}), and discussions of uncertainty are very spurious in the literature.
With the increasing success of recently proposed techniques by which certain aspects of the traditional comparative method can be automated, the formal representation of words, morphemes, cognate sets, and proto-forms has also improved. This makes it possible to address the problems of both the representation and the computation of uncertainty. 
Supervised approaches have led to major advances in automated phonological reconstruction; scholars provide a partially annotated dataset in which a certain number of proto-forms are already provided, and an algorithm is then trained on the data in order to propose new proto-forms for so far unobserved cognate sets. This task is very similar to the task of cognate reflex prediction, in which the word forms which have to be predicted are not proto-forms, but word forms from descendant languages, and algorithms have to predict the reflex of a cognate set in a given language based on the sound correspondence patterns and the reflexes of the cognate set in related languages. 
In the past decade, scholars have proposed quite a few new methods for both cognate reflex prediction and supervised phonological reconstruction. 
 
\citet{Meloni2021} tested recurrent neural networks on a dataset of Romance languages originally compiled by \citet{Ciobanu2014}, reporting very promising results on supervised approaches. This study was later extended by \citet{Kim2023}, who used a Transformer architecture \citep{Vaswani2017} and additionally tested the approach on a dataset of Chinese dialect varieties.
\citet{List2022d} proposed a new framework based on support vector machines to predict proto-forms from phonetic alignments, which they tested on six different datasets covering several different language families. In a recently organized Shared Task on cognate reflex prediction \citep{List2022f}, \citet{Kirov2022} proposed two methods that outperformed alternative approaches, one originally designed for the handling of place name pronunciations in Japan \citep{Jones2022XXX}, and one designed for the restoration of digital images in which pixels are missing \citep{Liu2018}. All in all, the most successful methods in the shared task all showed good performance: when retaining 90\% of the data for training, the methods differed on average by one sound from the attested word forms. 
 
While the task of unsupervised phonological reconstruction, where algorithms would reconstruct a proto-language from cognate sets from scratch, has not been sufficiently investigated so far (an early approach by \citealt{Bouchard-Cote2013} was only tested on Austronesian languages with the code never published), we can see that phonological reconstruction in a supervised setting has become a real option and could be integrated into computer-assisted workflows, in which scholars first annotate parts of their data, then compute new reconstructions automatically, and later refine them again.

With respect to the representation of uncertainty in reconstruction, linguists typically adopt ad-hoc solutions for individual language families or individual enterprises. In Indo-European studies, scholars express their uncertainty with respect to the three laryngeals (\textit{*h\textsubscript{1}}, \textit{*h\textsubscript{2}} or \textit{h\textsubscript{3}}) by writing a capital \textit{*H}. In their reconstruction of Old Chinese, \citet{Baxter2014} 
employ a complex notation system that puts uncertain parts of their reconstruction into brackets (with \textit{-[n]} meaning, for example, that the reconstruction could be either the final \textit{-n} or to \textit{-r}). In other cases, scholars mention alternative reconstructions only in comments. 
While both manual and automated methods are inherently fuzzy with respect to phonological reconstruction, so far, few methods have explicitly embraced fuzziness, trying to present uncertainty in reconstructions explicitly. An exception was the method of \citet{List2019a}, which offered degrees of uncertainty in the imputation of missing sounds in aligned cognate sets, but the fuzzy reconstructions were not further evaluated or inspected. 

\begin{table*}[tb]
\centering
\includegraphics[width=1.35\columnwidth]{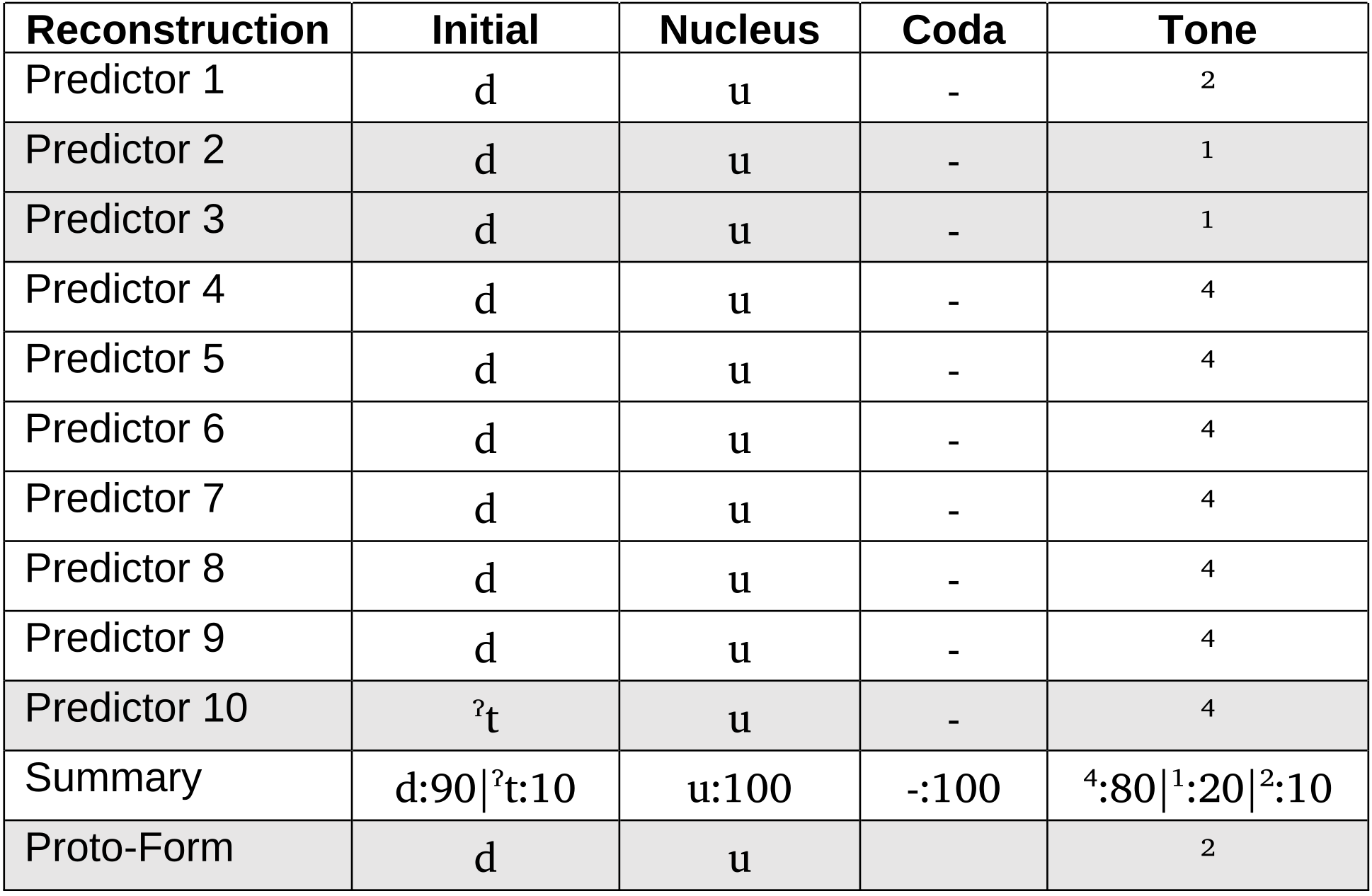}
\caption{Predictions for ``belly'' (cognate set 80) in Burmish. The table contrasts predicted word forms for all 10 different predictors, along with the aggregated representation (row Summary) and
contrasted with the reconstruction provided by the experts (Proto-Form).}
\label{tab:1}
\end{table*}

\section{Materials}
We work with three etymological datasets which are coded in Cross-Linguistic Data Formats (\url{https://cldf.clld.org}, \citealt{Forkel2018a,Forkel2020}), following the Lexibank workflow for the handling of multilingual wordlists \citep{List2022e}. 
The Burmish dataset consists of 8 Burmish languages, and 269 etymologies that are reflected in at least two descendant languages with a total of 1,442 reflexes. The data was originally compiled by \citet{Gong2020} and later converted to CLDF formats by \citet{List2022c} and further refined for the present study. It is accessible online at \url{https://github.com/lexibank/hillburmish}.
The Karen dataset consists of 10 Karenic languages, and offers 365 etymologies originally proposed by \citet{Luangthongkum2019}, which are reflected in at least 2 languages with a total of 2,866 reflexes. The data was also compiled for the study by \citet{List2022c} and slightly refined for this study. It is accessible online at \url{https://github.com/lexibank/luangthongkumkaren}.
The Panoan dataset consists of 20 Panoan languages, and includes the reconstruction of 514 cognate sets across 470 concepts proposed by \citet{Oliveira2014}. In total, the dataset features 7,305 reflexes. During the digitization of this dataset, all cognate sets were manually aligned \cite{blum2023opp}. It is accessible online at \url{https://github.com/pano-tacanan-history/oliveiraprotopanoan}.
\section{Methods}

\subsection{Representing Fuzzy Reconstructions}
We follow \citet{Bodt2022} who represent multiple options for the prediction of an individual sound by using the pipe symbol \texttt{|} as a separator for the different options. The symbol is often used in the meaning of ``or'' in regular expressions, which makes it particularly apt to represent uncertainty, since we can interpret a fictitious proto-form like \mbox{\textipa{[p a|i t]}} as a kind of a regular expression that matches both the form \mbox{\textipa{[p a t]}} and \mbox{\textipa{[p i t]}}. Note that this notation needs to be used with some care when more than one sound is treated as uncertain, since the resulting expression will always match the Cartesian product of the uncertain sounds. Thus, a fictitious proto-form \mbox{\textipa{[p a|i t|d]}} would match four distinct proto-forms, namely the forms \mbox{\textipa{[p a t]}}, \mbox{\textipa{[p i t]}}, \mbox{\textipa{[p a d]}}, and \mbox{\textipa{[p i d]}}. If scholars want to explicitly propose two different proto-forms only, e.g. \mbox{\textipa{[p a t]}} vs. \mbox{\textipa{[p i d]}}, our notation cannot be used. We recommend instead to assume two distinct forms, which can both be proposed as possible proto-forms for a given cognate set. Our fuzzy notation is thus only reserved for cases where the uncertainty is independent of contextual information that could be derived from the proto-form.

\begin{figure*}[tb]
    \tabular{cc}
    \includegraphics[height=3.8cm]{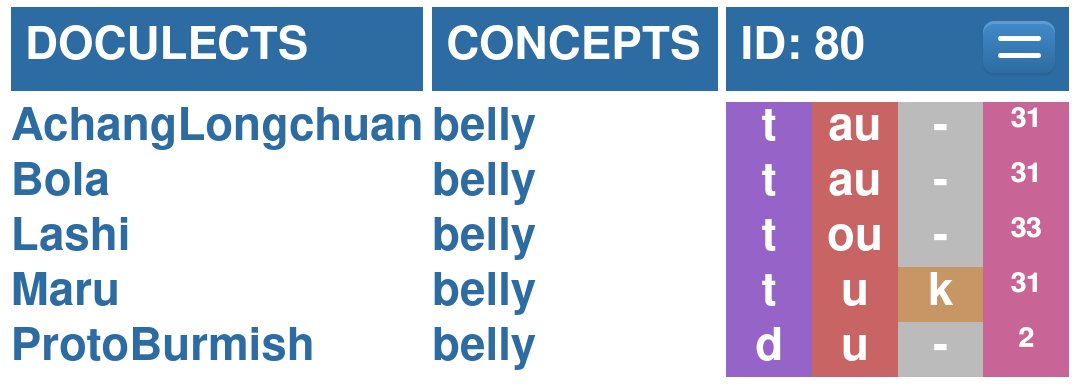} & 
    \includegraphics[height=3.8cm]{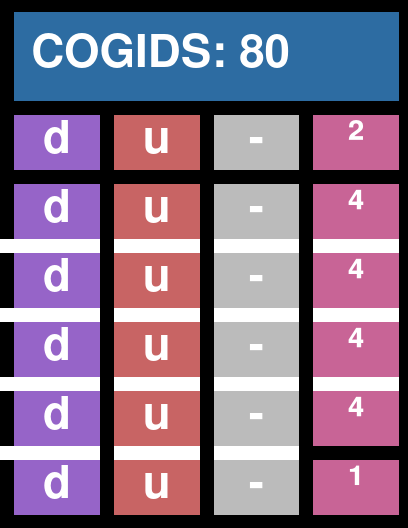} \\
    (a) Phonetic alignment of all word forms (including the proto-form).
    & (b) Quintile representation.\\
    \endtabular
    \caption{Contrasting the alignment representation with the quintile representation of the fuzzy reconstruction in the EDICTOR tool.}
\label{fig:1}
\end{figure*}

\subsection{Computing Fuzzy Reconstructions}

Our method for the creation of fuzzy reconstructions is straightforward. We expand the framework for supervised phonological reconstruction proposed by \citet{List2022d}, by drawing several samples from the same data, in which different parts of the forms are intentionally ignored. While the framework of \citeauthor{List2022d} starts from a training set in which proto-forms are provided and then a model is trained that can be used to predict proto-forms for data that has not been seen before, we draw multiple samples, drop a certain number of words from each sample, and use the method by \citeauthor{List2022d} to train the ``classifier'' that can be used to predict proto-forms from aligned cognate sets. Since we drop data in each of the samples, each sample will produce slightly different proto-forms, depending on the data which has been randomly ignored. The different proto-forms offered may point to problems in the original data, or reveal cognate sets that in fact underspecify the proto-form. 

While fuzziness could of course also be directly computed from the direct output of most approaches to supervised phonological reconstruction (since most of them work in a probabilistic manner that allows one to return not only the one and only best result, but also a certain range of candidates, see also \citealt{Fourrier2021}), our approach of using subsets of the original data has the clear advantage that it does not take the correctness of the original data for granted. When taking all data at once, it is difficult to spot irregularities in the data itself. When taking subsets, however, we test the robustness of the reconstructions for individual cognate sets. If the reconstruction, for example, depends on only one reflex, but this reflex is then discarded due to the subsampling in one particular run, the resulting reconstruction may turn out to be different, and this particular difference would then be accounted for in this trial and surface as an uncertainty. 

In the default settings of our method, we create 10 proto-form predictors from the annotated data and remove 10\% of the word forms in each of the samples. When creating an individual reconstruction, we feed our method with a concrete cognates set and then use all 10 predictors to predict proto-forms. The predictions are then summarized, and we count for each position in the original alignment how often which proto-sound occurs. These fuzzy reconstructions are then represented in the form of a sequence in which a column of the alignment is represented by at least one sound, and each possible sound is provided with the frequency in which it occurs in our 10 samples. Table 1 provides an example from the Burmish data for the fuzzy prediction procedure and the specific output produced by our method.

Since certain irregularities in the input data may be filtered from the different samples, irregular patterns which could lead an algorithm to propose erroneous proto-forms will be filtered out, and in this way the overall robustness of individual reconstruction can be tested.
\subsection{Visualizing Fuzzy Reconstructions}
Apart from the technical representation shown above, we have experimented with different ways to represent uncertainty or ``fuzziness'' in the tools we use to annotate etymological data. Since the manual curation of the cognate sets was carried out with the help of EDICTOR
(\citealt{EDICTOR}, 
\url{https://digling.org/edictor}), a web-based tool for the creation and curation of etymological datasets, we extended the EDICTOR representation of phonetic alignments by adding a representation which we call quintile-representation. In this representation, we represent the frequencies observed in the ten predictions with the help of a table with five rows, in which each row represents the attested symbols (converted from 10 to 5, to keep the table representation neat). This is shown in Figure \ref{fig:1}.

\begin{table*}[tb]
\centering
\tabular{|llccc|}
\hline
\bfseries Dataset   & 
\bfseries Prediction   &   
\bfseries Count &   
\bfseries Proportion &   
\bfseries Alignment Size \\ \hline\hline
\multirow{4}{2cm}{Burmish}   & correct      &     154 &         0.57 &             4.13\\
   & false        &     115 &         0.43 &             4.29\\
   & certain      &     199 &         0.74 &             4.13\\
   & uncertain    &      70 &         0.26 &             4.39\\\hline\hline
\multirow{4}{2cm}{Karen}     & correct      &     246 &         0.65 &             4.03 \\
     & false        &     133 &         0.35 &             4.27\\
     & certain      &     310 &         0.82 &             4.05\\
     & uncertain    &      69 &         0.18 &             4.41 \\\hline\hline

\multirow{4}{2cm}{Panoan}    & correct      &     405 &         0.79 &             4.25\\
    & false        &     109 &         0.21 &             5.14\\
    & certain      &     465 &         0.9  &             4.37\\
    & uncertain    &      49 &         0.1  &             5.14\\
\hline
\endtabular
\caption{Summary scores for the Burmish, Karenic, and Panoan predictions. Correct predictions refer to all those predictions which are identical with the reconstruction in the gold standard and which show no uncertainty. False predictions are those which show uncertainty or which are not identical with the proposed predictions in the gold standard. Certain predictions are those in which all ten trials on differently distorted data show the same results for a given proto-form. Uncertain predictions are those, accordingly, in which we observe differences. Alignment size refers to the size of the alignment of the cognate sets (conducted automatically).}
\label{tab:3}
\end{table*}

\subsection{Implementation}
The method of fuzzy reconstruction is implemented as Version 1.4.1 of the LingRex software package (\url{https://pypi.org/project/lingrex}, \citealt{LingRex}, which is itself an extension of the LingPy software package for quantitative tasks in historical linguistics (\url{https://pypi.org/project/lingpy}, \citealt{LingPy}). The quintile visualization is implemented as part of Version 2.2 of the EDICTOR tool (\url{https://digling.org/edictor}, \citealt{EDICTOR}).
The supplementary material shows how the package can be used and applied to the data, it is curated on GitHub (\url{https://github.com/lingpy/fuzzy/releases/tag/v1.0}) and has been archived with Zenodo (\url{https://doi.org/10.5281/zenodo.10007475}).  

\section{Evaluation}
Since we do not have a clear account on what constitutes a good ``fuzzy reconstruction'' and what constitutes a bad one, we closely analyzed the fuzzy reconstructions proposed for the three datasets and further investigated the results both quantitatively and qualitatively. In the following, we will thus report on the proportion of fuzzy reconstructions per datasets, the most frequently confused sounds in fuzzy reconstructions, and then report on major problems in the original data revealed through a close inspection of the fuzzy reconstructions proposed for the Burmish dataset.
\subsection{Proportion of Fuzzy Reconstructions}
As a first test of our approach, we computed fuzzy reconstructions from the three datasets and then compared whether~(a) the reconstructions were fuzzy at all, and~(b) to what extent they diverged from the proposed reconstructions. We explicitly chose a setting where we train and test the method on the same dataset, since we were not interested in the evaluation of the reconstruction method (which performs fairly well, but not perfect) but in the degree to which conclusions were based on the data in its entirety or different parts of it.
For each proto-form in the three datasets, we computed fuzzy reconstructions, from which we created consensus reconstructions using the notation shown in Table \ref{tab:1}. For each proposed reconstruction we tested~(a) if the reconstruction had conflicts (i.e. if it was ``fuzzy''), and~(b)~if it was not fuzzy, if it coincided with the reconstruction proposed by the linguist.

For the Burmish data, consisting of a total of 269 reconstructions, we arrived at the results reported in Table \ref{tab:3} (top). As can be seen from the table, the proportion of words reconstructed correctly by the approach and proportion of words that were reconstructed as ``certain'' (with no variation) is much larger than the proportion of false or uncertain reconstructions. Since a correct reconstruction has to be certain, it is not surprising that these numbers are similar, but the small difference of 57\% vs. 74\% shows that only a small part of the reconstructions identified as ``certain'' are also wrong. We find a small difference with respect to the alignment size (the number of words of which alignments for individual proto-forms are reconstructed) between correctly and falsely reconstructed proto-form, but due to the restricted syllable structure of Burmish languages, we do not find huge differences here. Additional studies are needed to find out what influences the certainty of automated reconstructions.

\begin{table*}[tb]
\centering
\tabular{|c|c|c|c|c|c|c|c|c|}
\hline
\multicolumn{3}{|c}{\bfseries Burmish} &
\multicolumn{3}{|c|}{\bfseries Karen} &
\multicolumn{3}{c|}{\bfseries Panoan} \\\hline
\bfseries Sound A & \bfseries Sound B & \bfseries Freq. &
\bfseries Sound A & \bfseries Sound B & \bfseries Freq. &
\bfseries Sound A & \bfseries Sound B & \bfseries Freq. \\\hline\hline
\textsuperscript{4}   & \textsuperscript{1} & 14 & \textipa{n}                   & \textipa{\r*n} & 18 & n  & \textipa{r{\super n}} & 24 \\\hline
\textsuperscript{4}   & \textsuperscript{3} & 9  & \textipa{n}                   & \textipa{\;N} & 14 & k  & -  & 13 \\\hline
\textipa{i}           & \textipa{e}         & 8  & \textipa{\;N}                   & \textipa{N} & 10 & \textipa{r{\super n}} & \textasciitilde  & 10 \\\hline
\textipa{N}           & -                   & 7  & \textsuperscript{55}                  & \textsuperscript{0} & 9  & -  & \textipa{t\super r} & 10 \\\hline
\textsuperscript{2}    & \textsuperscript{3} & 7  & \textipa{l}                   & \textipa{\r*l} & 8  & \textipa{n}  & -  & 9 \\\hline
-                     & \textipa{P}         & 7  & \textipa{\r*m}                   & \textipa{m} & 6  & \textipa{r{\super n}} & -  & 6 \\\hline
\textipa{{\super P}s} & \textipa{s}         & 6  & -                   &\textipa{P\textcorner} & 6  & \textipa{k}  & \textipa{t{\super r}} & 6 \\\hline
\textipa{{\super P}k} & \textipa{g}         & 6  & \textsuperscript{1} & \textsuperscript{0} & 5  & \textipa{t}  & \textipa{t{\super r}} & 5 \\\hline
\textsuperscript{2}   & \textsuperscript{4} & 6  & \textipa{k}                   & \textipa{g} & 5  & t  & -  & 5 \\\hline
\textipa{r}           & \textipa{j}         & 6  & \textipa{P\textcorner}                   & \textipa{P} & 4  & \textipa{r\super n} & r  & 5 \\\hline

\hline\endtabular

    \caption{Frequently confused sounds in the three datasets. Frequency refers to the number of cognate sets in which the automated reconstruction proposed alternative proto-sounds.}
    \label{tab:4}
\end{table*}

The results for the Karen data, consisting of 365 cognate sets, are shown in Table \ref{tab:3} (middle). As can be seen here, the number of correctly reconstructed proto-forms as well as the number for certain proto-forms are both higher than in the case of the Burmish data. One factor which may have contributed to this is that exceptional reflexes in this dataset have been manually identified and marked as such (as part of ongoing research), which means that certain irregularities in the data did not negatively impact the predictions. In contrast to the Burmish data, the differences in alignment size for correct vs. false proto-forms and certain vs. uncertain ones are more pronounced in this dataset. 

The results for the Panoan data in Table \ref{tab:3} show some interesting differences. Of the total of 514 cognate sets, 405 (79$\%$) are reconstructed correctly, a considerably higher number than for the other datasets. The number of reconstructions that are provided as ``certain'' is also higher (90$\%$) than for the other datasets. There is also a considerable difference in alignment size: The alignment size for correct (4.25) and ``certain'' (4.37) reconstructions is much lower than for false (5.14) and ``uncertain'' (5.14) reconstructions. Here, a larger alignment size arises as a possible source influencing the correctness and certainty of automated reconstructions. This illustrates that it may be worthwhile to investigate more closely how the reconstructions differ between different language families and between alignments of different sizes within the language families.

\subsection{Frequently Confused Sounds}
Each fuzzy reconstruction proposes at least two alternative sounds for one proto-segment in a given proto-form. Investigating these more closely in order to understand which sounds are frequently confused by the analysis, allows us to gain insights into those sounds in the proto-languages which are particularly difficult to reconstruct. Table \ref{tab:4} provides the 10 most frequently confused sound pairs in both datasets (our workflow reports all of them).

\begin{table*}[tb!]
\centering
\includegraphics[width=\textwidth]{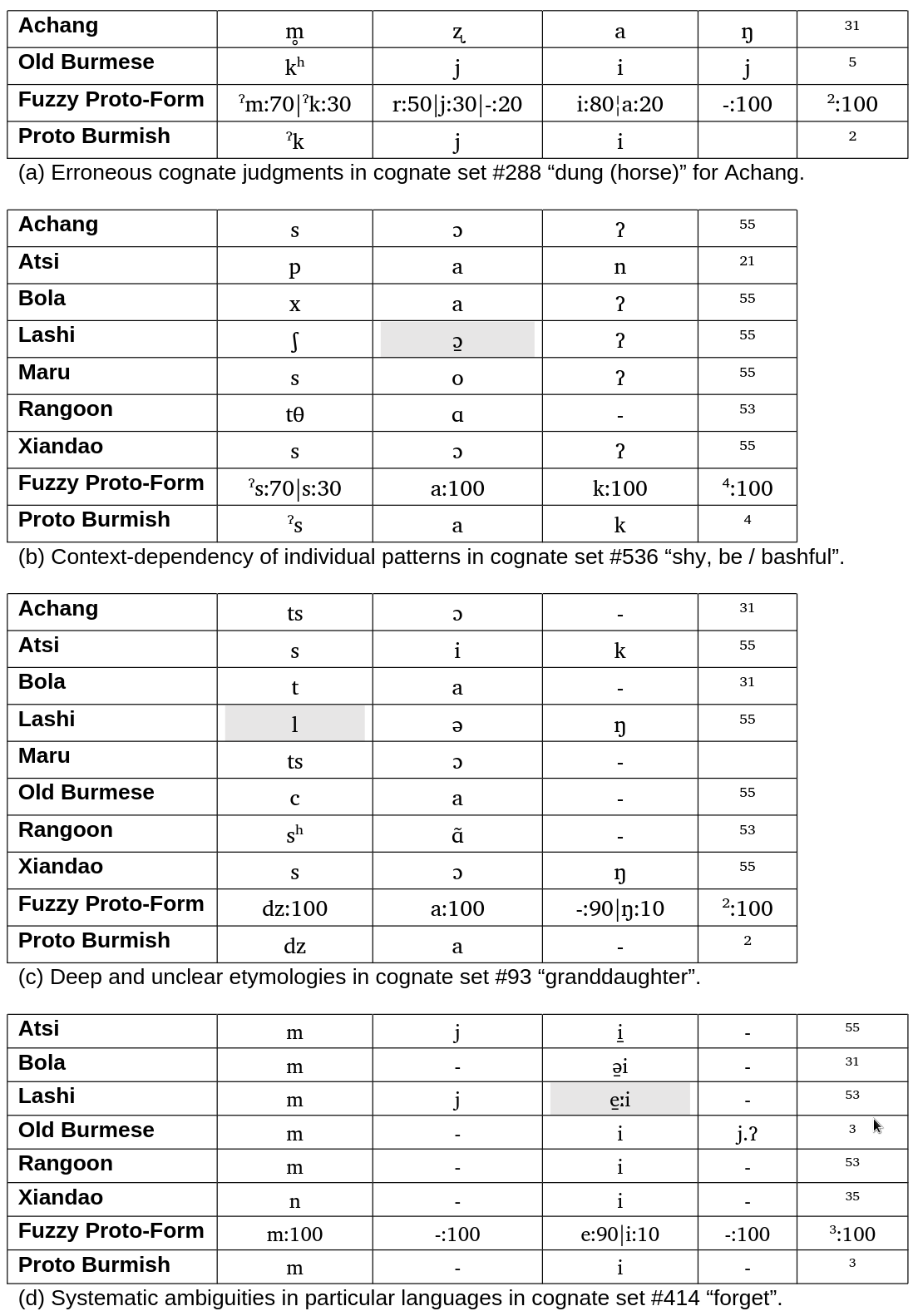}

\caption{Examples for causes of fuzziness in Burmish reconstructions. }
\label{tab:5-8}
\end{table*}

As can be seen from the individual results for the Karen and Burmish data, the particular problems are quite different across both datasets and cannot be directly compared with each other. A major difficulty in the Karenic data is the reconstruction of voiceless sonorants 
(\textipa{[\textsubring{n}]}, \textipa{[\textsubring{l}]}, \textipa{[\textsubring{m}]}, etc.), which the author proposes on the basis of the tonal development in some of the descendant languages \citep{Luangthongkum2019}. 
Since there are quite a few exceptions with respect to the tonal development, we find that the original reconstruction itself cannot always indicate clearly whether a proto-sound should be voiced or voiceless, which is at times marked by putting the h, which is used to mark a sonorant as voiceless in parentheses (resulting in forms like \textipa{(h)n-}, ibid.). The confusion of the tone marked as [\textsuperscript{0}] with other tones results from our annotation practice of certain weak syllables, in which originally no tone was reconstructed. Since we wanted to indicate a tone nevertheless, to fill the slot in our alignment, the [\textsuperscript{0}] thus marks an underspecified value, which -- as the fuzzy reconstructions show -- might just as well be given a more concrete reconstruction. 

In the Burmish data, on the other hand, we find three major types of confusion. The first relates to the reconstruction of tones. The reconstruction here is often predicted by the nature of the final consonants, which are not actively used in the automated reconstruction method. This may explain the confusion in this case. The second case relates to the reconstruction of gaps (marked by the symbol \textipa{[-]}), which are often confused with sounds occurring in coda position, such as 
\textipa{[N]}, \textipa{[r]}, or \textipa{[\textglotstop]}. The confusion of pre-glottalized initials like \textipa{[\textsuperscript{\textglotstop}s]} and \textipa{[\textsuperscript{\textglotstop}k]} and their non-glottalized counterparts also results from the fact that the reconstruction of pre-glottalized initials depends on the vowels that appear as reflexes in certain Burmish languages. Since this information was not taken into account by our automated method, it is not surprising that results may vary here. 
The confusion resulting from information that is not represented in the individual column of an alignment but in other parts shows that additional analyses in which we take the vowel information in the Burmish languages and the tonal information in the Karenic languages into account would be useful in the future.

The confused sounds for the Panoan data fall mainly into two groups,  (a)~word-final stops \textipa{[k]}, \textipa{[t]}, and \textipa{t\super r]}, and (b)~word-final nasal and liquid consonants 
\textipa{[n]}, \textipa{[r\super n]}, and \textipa{[r]}. Interestingly, those cases are either described as uncertain due to missing data by the original author (word-final nasals), or are the most debated feature of the reconstruction (word-final stops instead of three-syllabic words with an open syllable). The word-final sonorants are described by the author of the dataset as being uncertain due to the lack of reflexes in Kaxarar\'i, a nearly undescribed Panoan language which retains the contrast of word-final \textipa{[r]} and \textipa{[n]}. This is the main source of confusion for \textipa{[n]}, \textipa{[r\super n]}, and \textipa{[r]}, but also for some of the word-final stops. Here, the confusion primarily arises because the reconstructions are proposed based on reflexes of only a few languages, which often do not provide sufficient evidence for identifying the phonemic nature of the reflex in the proto-language. Our method thus correctly identifies the cases in which the provided reconstructions should be considered ``fuzzy'', given their uncertain nature. It also validates the large part of correct reconstructions.

\subsection{Detailed Examples for Burmish}
A closer inspection of discrepancies in the Burmish data reveals four major kinds of problems, namely (1)~problems resulting from problematic cognate judgments in our data, (2)~problems resulting from the context-dependency of reconstructions which our automated reconstruction method does not (yet) account for, (3)~problems resulting from deep etymologies which are not (yet) well understood, and (4)~problems resulting from some systematic and so far not clearly understood ambiguities in particular languages.

\subsubsection{Problematic Cognate Judgments}
The method allows us to identify quite a few cases where individual cognate judgments turned out to be erroneous and should be modified in future versions of our data. As an example, consider cognate set \#288 ``dung (horse)'' in our Burmish data, shown in Table \ref{tab:5-8}~(a). That erroneous cognate judgments occur in larger etymological projects is inevitable to some degree. Here, our method for the reconstruction of ``fuzzy'' proto-forms directly helps us to identify and eliminate these problems in future releases of our data.

\subsubsection{Context-Dependency of Reconstructions}
While phonological reconstruction can, in the majority of the cases, be successfully carried out by considering individual correspondence patterns alone, there are certain cases where it is not enough to look at a pattern in isolation. What needs to be done instead is to evaluate the pattern in combination with other patterns from the same alignment. Although our method for automatic phonological reconstruction was designed in such a way that it can in theory account for this context-dependency of individual reconstructions, we did not take specific and known processes of sound change in the Burmish and the Karenic data into account, when applying our method to the data. This was done intentionally, since we wanted to see how far we can get with a unified approach. Individual reconstruction errors and cases of uncertainty in the automated reconstruction, however, show that context-dependency should be accounted for in future applications of our approach.

As an example for the problems resulting from ignoring context-dependencies, Table \ref{tab:5-8}~(b) shows the reconstruction for the cognate set \#536 ``shy, be / bashful'' in the Burmish data. As we can see, Lashi has a tense vowel (indicated by the bar under the vowel, shaded in gray in the table). Tense vowels are taken as evidence for the reconstruction of pre-glottalized initials in Proto-Burmish, while the correspondence pattern of the initial itself does not provide concrete evidence for the presence or absence of pre-glottalization. As a result, we can see that the automated method is uncertain, proposing a pre-glottalized initial in 70\% of the cases, and a plain initial in 30\%.

\citet{List2022c} have described in detail, how context-dependency can be accounted for by means of
``extended alignments'' or ``multi-tiered sequence representations''.
Future studies are needed to test how well these work to handle the Burmish (and also the Karenic) data.

\subsubsection{Etymologies with Unclear Variation}

There are a couple of cognate sets where we have in principle no doubts that the words in question are cognates, but we have problems to understand the etymological processes in full. These deep etymologies with unclear variation are usually of great importance when it comes to advancing existing reconstruction systems. However, since they may well reflect processes predating the history of the language family in question, the solution may only be achieved when taking more languages from higher clades of the language family in question into account.

As an example, consider Table \ref{tab:5-8}~(c), showing alignments and reconstructions for cognate set \#93 ``granddaughter''. While it is possible that all forms are cognate, it is hard to decide for sure, given that individual languages show reflexes which do not follow our expectations. Thus the initial \textipa{[l-]} in Lashi does not fit the pattern at all, and from the pattern, we have evidence for three different finals in the data. Future work may either show that we have to refine the cognate assignment of individual works in this pattern, or we may find solutions in certain etymological processes that counteract regular sound change.

\subsubsection{Systematic Ambiguities in Languages}

As a final type of difficulty, there are cases where we have clear ambiguities in individual languages which we cannot (yet) resolve and explain. As an example, Table \ref{tab:5-8}~(d) shows ambiguities for the reconstruction of the nucleus vowel in the cognate set \#414 ``forget'', where our reconstruction proposes \textit{*i}, while the automated method sees more evidence for \textit{*e} (90\%) and less evidence for \textit{*i} (10\%). The evidence from the correspondence pattern is difficult to interpret. While Old Burmese points to an \textit{*i}, Bola and Lashi point to an \textit{*e}. The fuzzy reconstruction approach thus correctly points to the ambiguity of the pattern in the light of our data.

\section{Conclusion}

In this study, we have introduced some novel ideas regarding the handling of uncertainty in phonological reconstruction in historical linguistics. We have tried to 
show that it may be useful to transparently record uncertainty not only in classical reconstructions but also in reconstructions proposed by automated approaches. 
 
These considerations resulted in the proposal of a new framework for fuzzy reconstructions that allows one to compute fuzzy reconstructions from annotated comparative wordlists. Applying this framework to three datasets, two from the Sino-Tibetan language family (Burmish and Karenic), as well as on the Panoan language family, we have shown how the framework can be used to compute the degree of uncertainty in a given dataset, how frequently confused sounds can be computed, and how an individual inspection of the data reveals major classes of errors in the original data.
 
In the future, we hope to refine our current approach in three ways. First, we want to enhance the individual automatic reconstructions for the Burmish data and the Karenic data by taking the context of important sounds into account. Second, we want to enhance our data by correcting cases where we identified problematic cognate judgments. Third, we want to apply our method to more data from other language families in order to see how the approach performs there.

\section*{Supplementary Material}
Supplementary material needed to replicate the experiments shown here, including data and code, has been curated on GitHub (\url{https://github.com/lingpy/fuzzy/releases/tag/v1.0}) and archived with Zenodo (\url{https://doi.org/10.5281/zenodo.10007475}).

\section*{Limitations}
Our approach comes with some limitations. First, since the computation depends on the original data, fuzzy cognates do not only reflect true cases of uncertainty (where scholars would assess that the evidence is not enough to decide for one particular among several sounds) but can also be due to errors in the originally coded data. Second, since we use a specific procedure of grouping sounds in those cases where a proto-sound does not correspond to any sound in the descendant data,\footnote{This is labelled \emph{trimming} in \citealt{List2022d}, but the term does not seem a good choice, given that trimming in biology refers to cases where entire columns in an alignment are dropped, see \citealt{blum-list-2023-trimming}.} our automated reconstruction approach currently may reconstruct phonotactically incorrect proto-forms. These forms may consist, for example, of two identical finals. Third, as also discussed in the study, context-dependencies which are not explicitly handled in the reconstruction procedure may yield ambiguities even in those cases, where we know they should not occur. Fourth, so far, our experiments have only been dealing with alignments that were computed automatically. Manually annotated alignments have not yet been tested.

\section*{Ethics Statement}
Our data are taken from publicly available sources.
There are no ethical issues or conflicts of interest
in this work that we would know of at the time of writing.

\section*{Acknowledgements}
This project was supported by the ERC Consolidator Grant ProduSemy (PI Johann-Mattis List, Grant No. 101044282, see \url{https://doi.org/10.3030/101044282}) and the Max Planck Research Grant CALC³ (\url{https://calc.digling.org}, PI Johann-Mattis List). Views and opinions expressed are however those of the author(s) only and do not necessarily reflect those of the European Union or the European Research Council Executive Agency (nor any other funding agencies involved). Neither the European Union nor the granting authority can be held responsible for them. We thank the anonymous reviewers for helpful comments and all people who share their data openly, so we can use it in our research.


\end{document}